%% file: egpaper_final.tex
\documentclass[10pt,twocolumn,letterpaper]{article}

\usepackage{cvpr}
\usepackage{times}
\usepackage{epsfig}
\usepackage{graphicx}
\usepackage{amsmath}
\usepackage{amssymb}
\usepackage{caption}

\usepackage{graphicx}
\usepackage{subfigure}
\usepackage{threeparttable}
\usepackage[marginal, flushmargin]{footmisc}
\usepackage{algorithm}
\usepackage{algorithmic}
\usepackage{multirow}
\usepackage{bm}

\usepackage[nottoc]{tocbibind}
\usepackage[square,numbers,sort]{natbib}
\usepackage[breaklinks=true,bookmarks=false]{hyperref}

\cvprfinalcopy 

\def\cvprPaperID{2412} 

\begin{document}

\title{A Compact DNN: Approaching GoogLeNet-Level Accuracy of \\Classification and Domain Adaptation}

\author{Chunpeng Wu\textsuperscript{1}\thanks{Part of this work was done while C. Wu was an intern at LG San Jose Lab.}, Wei Wen\textsuperscript{1}, Tariq Afzal\textsuperscript{2}, Yongmei Zhang\textsuperscript{2}, Yiran Chen\textsuperscript{3}, and Hai (Helen) Li\textsuperscript{3}\\
	\textsuperscript{1}Electrical and Computer Engineering Department, University of Pittsburgh, Pittsburgh, PA 15260\\
	\textsuperscript{2}LG San Jose Lab, Santa Clara, CA 95054\\
	\textsuperscript{3}Electrical and Computer Engineering Department, Duke University, Durham, NC 27708\\
	{\tt\small \{chunpeng.wu,wei.wen\}@pitt.edu, \{tariq.afzal,jenny.zhang\}@lge.com}\\
	{\tt\small \{yiran.chen,hai.li\}@duke.edu}
	}

\maketitle

\input{Abstract}
\input{Introduction}

\input{Related_work}
\input{Proposed_method}

\input{Experiments}

\input{Conclusion}

\input{Acknowledgement}
{\small
\bibliographystyle{ieee}
\bibliography{egbib}
}

\end{document}

%% file: Abstract.tex
\begin{abstract}

Recently, DNN model compression based on network architecture design, \textit{e.g.}, SqueezeNet, attracted a lot of attention.
Compared to well-known models, these extremely compact networks don't show any accuracy drop on image classification.
An emerging question, however, is whether these compression techniques hurt DNN's learning ability other than classifying images on a single dataset. 
Our preliminary experiment shows that these compression methods could degrade domain adaptation (DA) ability, though the classification performance is preserved.
In this work, we propose a new compact network architecture and unsupervised DA method.
The DNN is built on a new basic module Conv-M that provides more diverse feature extractors without significantly increasing parameters.
The unified framework of our DA method will simultaneously learn invariance across domains, reduce divergence of feature representations and adapt label prediction.
Our DNN has 4.1M parameters---only 6.7\% of AlexNet or 59\% of GoogLeNet.
Experiments show that our DNN obtains GoogLeNet-level accuracy both on classification and DA, and our DA method slightly outperforms previous competitive ones. 
Put all together, our DA strategy based on our DNN achieves state-of-the-art on sixteen of total eighteen DA tasks on popular Office-31 and Office-Caltech datasets.

\end{abstract}

%% file: Introduction.tex
\section{Introduction and Motivation}
\label{sec:introduction}

The success of deep neural networks (DNNs) encourages extensive applications on various types of platforms, \textit{e.g.}, self-driving cars and VR headsets. 
To overcome the hardware constraints, DNN model compression techniques, from learning based~\cite{Liu_CVPR15,Han_ICLR16,Wen_NIPS16} to network architecture design~\cite{Iandola_arxiv16,Wang_arxiv16,Paszke_arxiv16}, recently attracted a lot of attention.
Interestingly, most of these extremely compact DNN models do not show accuracy drop on image classification. 
A critical question emerges, however, other than classifying images on a single dataset, whether the compression methods hurt DNN's learning ability.

In this work, we attempt to bridge the gap between compressed DNN architecture and its domain adaptation (DA) ability.
The DA ability is to evaluate whether a machine learning model can capture the \textit{covariate shift}~\cite{Shimodaira_JSPI00} between source and target domains, and adapt itself to remove the divergence.  
A model with outstanding semi-supervised or unsupervised DA ability can greatly reduce the requirement of manually labeled examples for real-world applications.

\setlength{\tabcolsep}{5pt}
\begin{table*}
	\begin{center}
		\caption{Image classification and unsupervised DA accuracy of DNN models on Office-31 dataset.}
		\label{table:comp_classification_da}
		\small
		\begin{threeparttable}
			\begin{tabular}{l|c|c|ccc}
				\hline\noalign{\smallskip}
				& \multirow{3}{*}{\#Parameter} & \multirow{3}{*}{Classification} & Task1  & Task2  & Task3 \\
				&                              &                                 & AMAZON & DSLR   & WEBCAM \\
				&                              &                                 & WEBCAM & WEBCAM & DSLR   \\
				\hline\hline\noalign{\smallskip}
				AlexNet~\cite{Alex_NIPS12}                & 61 M    & 57.2           & 73.0   & 96.4    & \textbf{99.2}\\
				FaConvNet~\cite{Wang_arxiv16}             & 2.8 M   & 70.1           & 71.8    & 94.3   & 98.1\\
				SqueezeNet~\cite{Iandola_arxiv16}  & \textbf{1.2 M}   & 57.5         & 64.4    & 92.8   & 96.4\\
				\hline\noalign{\smallskip}
				Rev-FaConvNet                    & 4.8 M   & \textbf{70.3}  & \textbf{74.1}    & \textbf{96.5}   & \textbf{99.2}\\
				Rev-SqueezeNet                            & 2.2 M   & 57.9           & 66.9    & 93.9   & 98.8\\		
				\hline
			\end{tabular}
		\end{threeparttable}
	\end{center}
\end{table*}
\setlength{\tabcolsep}{5pt}
 
We observe DA accuracy degradation from model compression methods based on architecture design, \textit{e.g.}, a DNN with GoogLeNet-level~\cite{Szegedy_CVPR15} classification accuracy only obtains AlexNet-level~\cite{Alex_NIPS12} DA accuracy.
Table~\ref{table:comp_classification_da} shows our experimental results. 
SqueezeNet~\cite{Iandola_arxiv16} and FaConvNet~\cite{Wang_arxiv16} are used to compare with AlexNet as they are respectively the smallest DNN model achieving AlexNet-level and GoogLeNet-level accuracy on image classification, to our best knowledge.
The popular dataset ImageNet'12~\cite{Russakovsky_IJCV15} is adopted as image classification benchmark. 
Three standard DA tasks on Office-31~\cite{Saenko_ECCV10} dataset are adopted, and the unsupervised DA method used for all DNNs in Table~\ref{table:comp_classification_da} is GRL~\cite{Ganin_ICML15}. 
The DNNs are pre-trained on ImageNet'12, and then fine-tuned for all DA tasks. 
There is a big DA accuracy difference between AlexNet and SqueezeNet though the two networks have almost the same classification accuracy.
FaConvNet, which outperforms AlexNet by 12.9\% on classification, also slightly lags behind AlexNet on DA.

\begin{figure}
	\centering
	\includegraphics[width=7cm]{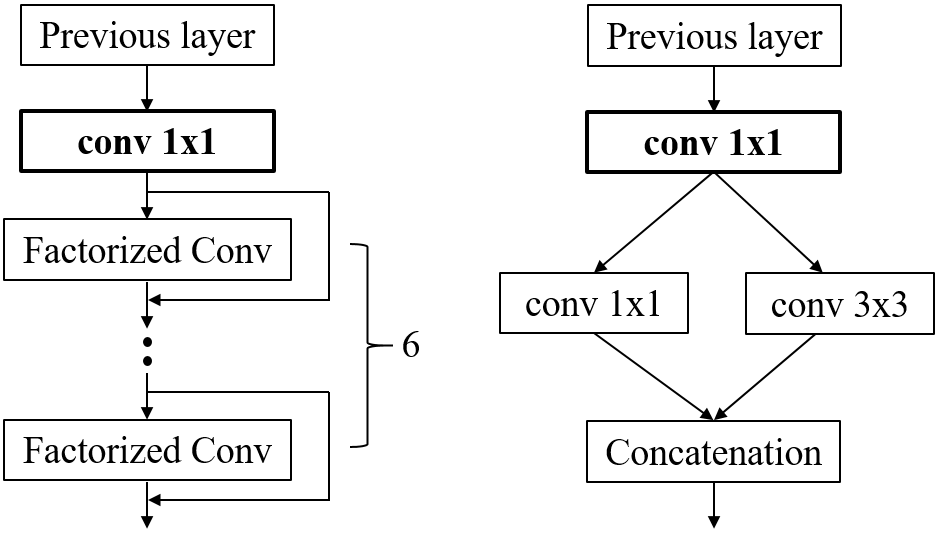}
	\caption{Basic modules adopted in FaConvNet~\cite{Wang_arxiv16} (left) and SqueezeNet~\cite{Iandola_arxiv16} (right). Both modules use the ``bottleneck'' layer as shown in bold.}
	\label{fig:module_sq_fa}
	\vspace{-8pt}
\end{figure}

Intuitively, increasing parameters will lead to better accuracy.
Our following experiment shows that the DA accuracy of SqueezeNet and FaConvNet can be improved, but can not reach the same level as their classification by solely boosting parameter numbers.
Specifically, without changing the structure of the two models, we increase the parameters of FaConvNet and SqueezeNet. 
The basic modules respectively adopted in FaConvNet and SqueezeNet are first compared, as shown in Figure~\ref{fig:module_sq_fa}. 
The shared feature of these two modules is the ``bottleneck'' layer \textit{conv 1}$\times$\textit{1} as denoted in bold. 
We hence gradually increase parameters of all ``bottleneck'' layers in FaConvNet and SqueezeNet until no DA accuracy benefit could be obtained. 
The parameters in other layers (\textit{e.g.}, the first convolutional layer in FaConvNet and SqueezeNet) are then increased until no accuracy gain.
The final DA accuracy of the adapted models \textit{Rev-FaConvNet} and \textit{Rev-SqueezeNet} are respectively shown in Table~\ref{table:comp_classification_da}.
Our expectation is that Rev-FaConvNet's accuracy can be much higher than AlexNet.
Rev-FaConvNet, however, only slightly outperforms AlexNet, with almost 70\% more parameters.

The objective of this work is to develop a compact DNN architecture which can achieve the same level accuracy on classification and DA.
Our solution offers four important features.
First, our DNN has 4.1M parameters, which is only 6.7\% of AlexNet or 59\% of GoogLeNet. 
The compactness of our network can be attributed to the use of a new module \textit{Conv-M} which is a parameter-saving module, while extract more details based on multi-scale convolution and deconvolution, inspired by GoogLeNet's \textit{Inception}. 
Second, our DA method consists of three components: Learning invariance across domains, reducing discrepancy of feature representations, and predicting labels.
Third, experiments show that our DNN obtains GoogLeNet-level accuracy both on classification and DA.
The DA accuracy gap between GoogLeNet and other compact DNNs (FaConvNet and Rev-FaConvNet) is much larger.
Fourth, the unified framework of our DA method slightly outperforms previous competitive methods, and our DA method based on our DNN network achieves state-of-the-art on sixteen of total eighteen DA tasks on the popular Office-31 and Office-Caltech~\cite{Gong_CVPR12} datasets.

%% file: Related_work.tex
\section{Related Work}
\label{sec:related_work}

\textbf{DNN model compression} with little accuracy drop on image classification traditionally are learning based.
Liu \textit{et al.}~\cite{Liu_CVPR15} zero out more than 90\% of AlexNet's parameters using a sparse decomposition, while Wen \textit{et al.}~\cite{Wen_NIPS16} regularize a DNN model with structured sparsity based on group Lasso. 
Han \textit{et al.}~\cite{Han_ICLR16} prune the small-weight connections and retrain the DNN with the remaining connections.
More recent research began to shrink a model directly based on network architecture design. 
SqueezeNet~\cite{Iandola_arxiv16} is built on the \textit{fire} module which feeds ``squeeze'' layer (1$\times$1 convoluton) into ``expand'' layer (a combination of 1$\times$1 and 3$\times$3 convoluton). 
The basic structure of FaConvNet~\cite{Wang_arxiv16} is \textit{Convolutional Layer as Stacked Single Basis Layer}.
A popular design methodology of compact architectures extensively uses small convolutional kernels (1$\times$1 and 3$\times$3), especially the linear projection as the \textit{conv} \textit{1}$\times$\textit{1} layer shown in bold in Figure~\ref{fig:module_sq_fa}. 
Based on the preliminary experimental result in Table~\ref{table:comp_classification_da}, we argue that it is necessary to redesign the basic module of these extremely shrunk DNNs, \textit{e.g.}, FaConvNet and SqueezeNet, by introducing more diverse operations of feature extraction, in order to achieve high accuracy on both classification and DA. 
The challenge lies in that more complex feature extraction methods, \textit{e.g.}, multi-scale convolution, often result in the steep increase of parameters, as the basic module will be used reapeatedly.
The shortcut connection used in ResNet~\cite{He_arxiv15}, for instance, can be understood as a parameter-saving solution of multi-scale feature integration. 
We will adopt methods other than this bypass structure. 

\textbf{Unsupervised DA.}
Following the early attempt of re-weighting samples from source domain~\cite{Huang_NIPS06}, Shekhar \textit{et al.}~\cite{Shekhar_CVPR13} learn dictionary based representations by minimizing the divergence between the source and target domains. 
The subspace based methods, on the other hand, evaluate the distance between domains in a low-dimensional manifold~\cite{Gong_CVPR12} or in terms of Frobenius norm~\cite{Fernando_ICCV13}.
DNN based methods have been proposed recently.
Glorot \textit{et al.}~\cite{Glorot_ICML11} and Chopra \textit{et al.}~\cite{Chopra_ICMLW13} learn cross-domain features using auto-encoders, followed by the label prediction.
A more popular strategy is to combine feature adaptation with label prediction as s unified framework.
DDC~\cite{Tzeng_arxiv14} introduces adaptation layers and domain confusion metric into a CNN architecture, while GRL~\cite{Ganin_ICML15} combines classifiers of label and domain using a gradient reverse layer.
DAN~\cite{Long_ICML15} and RTN~\cite{Long_NIPS16} focus on effectively measuring feature representations in kernel spaces.
TRANSDUCTION~\cite{Sener_NIPS16} jointly optimizes the target label and domain transformation parameters.
Our DA method adopts a unified framework, which can simultaneously learn invariance across domains, reduce divergence of feature representations and adapt label prediction.

\textbf{DNN based image segmentation.}
The DNNs of segmentation and classification mainly differ in the use of up-sampling layers to recover resolution.
Various up-scaling methods have been proposed and adopted, such as straightforward bicubic interpolation~\cite{Dong_arxiv15}, learning based deconvolution~\cite{Noh_ICCV13}, and unpooling~\cite{Hong_NIPS15,Badrinarayanan_arxiv15}.
We improve the deconvolution~\cite{Noh_ICCV13} to remove artifacts that will be described in Section~\ref{sec:our_dnn_arch}, and use it as a type of shape feature extractor in the basic module of our DNN.
With the consideration of training convergence speed, the unpooling with fewer parameters is a better choice, compared to deconvolution, especially for small-scale and medium-scale problems. 
So we adopt unpooling for sample reconstruction in our DA method.
In addition, different strategies have been presented to train segmentation networks.
SegNet-Basic~\cite{Badrinarayanan_arxiv15} is directly trained as a whole.
Long \textit{et al.}~\cite{Long_CVPR15}, on the other hand, adapt a popular classification network into a fully convolutional network (FCN), and fine-tune it for segmentation tasks.
Yu \textit{et al.}~\cite{Yu_ICLR16} show that accuracy can be further improved by plugging their context module into existing segmentation model.
Our decoder design for sample reconstruction is inspired by FCN, while our structure is simpler than the multi-stream structure in FCN.

\begin{figure}
	\centering
	\includegraphics[width=7.0cm]{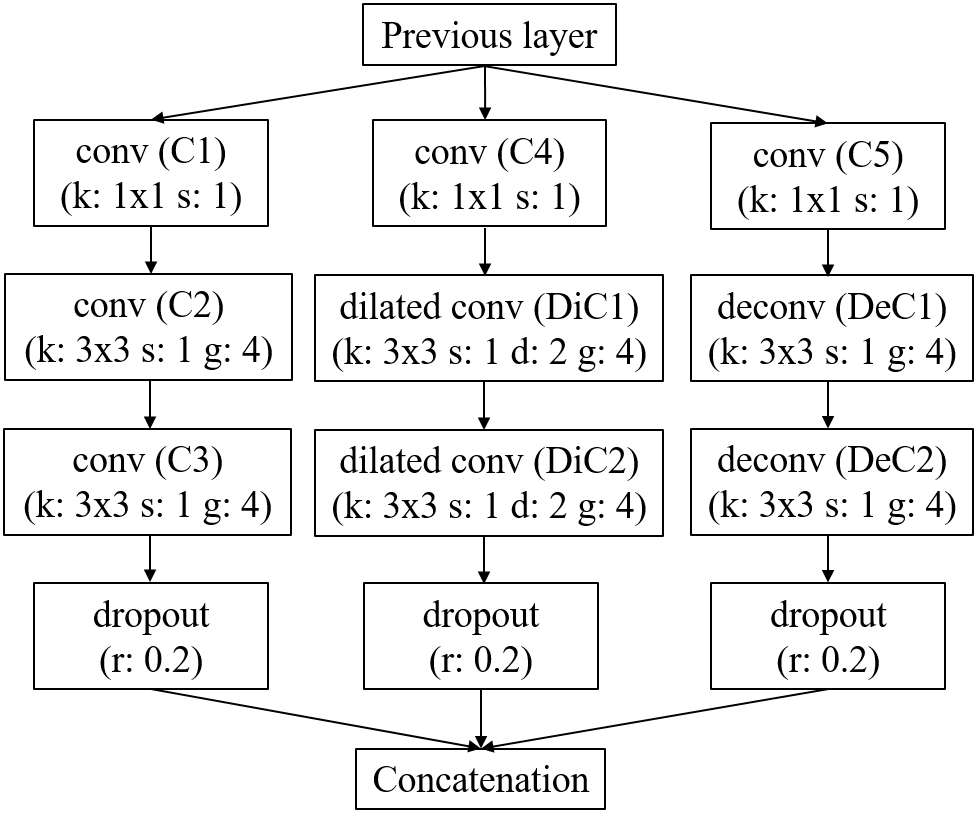}
	\caption{Module Conv-M used in our DNN. The output of \textit{deconv} is cropped to its input size. The ReLU is adopted for all types of convolution, which is not shown in the figure for simplicity.}
	\label{fig:module_Conv-M}
\end{figure}

\setlength{\tabcolsep}{3pt}
\begin{table*}
	\begin{center}
		\caption{Our DNN architecture (Basic parameter settings of the module Conv-M are shown in Figure~\ref{fig:module_Conv-M}).}
		\label{table:our_dnn_arch}
		\small
		\begin{threeparttable}
			\begin{tabular}{c|c|c|c|c|c|c|c|c|c|c|c|c|c}
				\hline\noalign{\smallskip}
				\multirow{2}{*}{Layer} & \multirow{2}{*}{Type/Module} 
				& \multirow{2}{*}{Output size}
				& Filter size/Stride
				& \multicolumn{9}{|c|}{\#Feature maps (Conv-M)}
				& \multirow{2}{*}{\#Parameters}\\
				\cline{5-13}
				& & & (If not Conv-M)  &  C1  &  C2  &  C3  &  C4  &  DiC1  &  DiC2  &  C5  &  DeC1  &  DeC2  &\\
				\hline\hline
				1 & input            & 224$\times$224$\times$3   & & & & & & & & & &\\
				\hline
				2 & convolution      & 224$\times$224$\times$64  & 7$\times$7/1 (x64)   & & & & & & & & &   & 9,408\\
				\hline
				3 & max-pooling      & 112$\times$112$\times$64  & 3$\times$3/2         & & & & & & & & & &\\
				\hline
				4 & Conv-M           & 112$\times$112$\times$160 &                      
				&  64  &  64  &  64    &  64   & 64    &  64    &  32  &  32  &  32  & 51,712\\
				\hline
				5 & max-pooling      & 56$\times$56$\times$160   & 3$\times$3/2         & & & & & & & & & &\\
				\hline
				6 & Conv-M           & 56$\times$56$\times$320   &
				& 128  & 128  & 128    &  128  & 128   & 128    &  64  &  64  &  64  & 217,088\\
				\hline
				7 & Conv-M           & 56$\times$56$\times$320   &
				& 128  & 128  & 128    &  128  & 128   & 128    &  64  &  64  &  64  & 268,288\\                       
				\hline
				8 & max-pooling      & 28$\times$28$\times$320   & 3$\times$3/2         & & & & & & & & & &\\
				\hline
				9 & Conv-M           & 28$\times$28$\times$576   &
				& 144  & 256  & 256    &  144  & 256   & 256    &  64  &  64  &  64  & 591,872\\
				\hline
				10 & Conv-M          & 28$\times$28$\times$576  &
				& 144  & 256  & 256    &  144  & 256   & 256    &  64  &  64  &  64  & 681,984\\
				\hline
				11 & max-pooling     & 14$\times$14$\times$576   & 3$\times$3/2         & & & & & & & & & &\\
				\hline
				12 & Conv-M          & 14$\times$14$\times$688   &
				& 160  & 256  & 280    &  160  & 256   & 280    &  64  &  128 &  128 & 783,360\\
				\hline
				13 & Conv-M          & 14$\times$14$\times$688   &
				& 160  & 256  & 280    &  160  & 256   & 280    &  64  &  128 &  128 & 826,368\\
				\hline
				14 & avg-pooling     & 1$\times$1$\times$688     & 14$\times$14/1       & & & & & & & & & &\\
				\hline
				15 & linear & 1$\times$1$\times$1000    & 1$\times$1/1 (x1000) & & & & & & & & &   & 688,000\\
				\hline  				
				& & & & & & & & & & & & & \textbf{4.1 M}\\
				\hline
			\end{tabular}		
		\end{threeparttable}
	\end{center}
\end{table*}
\setlength{\tabcolsep}{3pt}

\begin{figure}[ht]
	\centering
	\includegraphics[width=5.8cm]{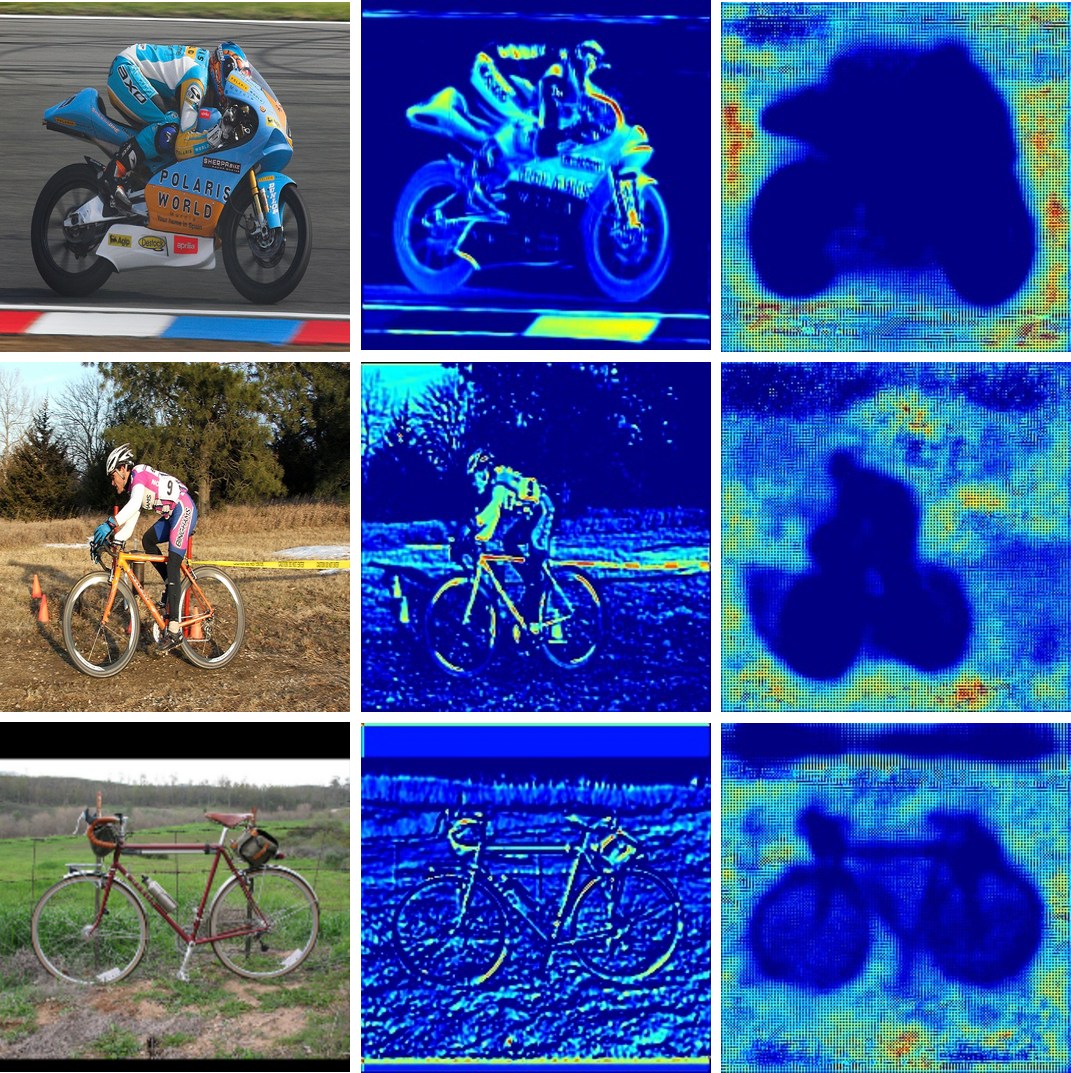}
	\caption{Visualization of activations in the same Conv-M module in our network: Convolution (middle) and deconvolution (right).}
	\label{fig:deconv_conv}
	\vspace{-10pt}
\end{figure}

\begin{figure*}
	\centering
	\vspace{-8pt}
	\includegraphics[width=17.5cm]{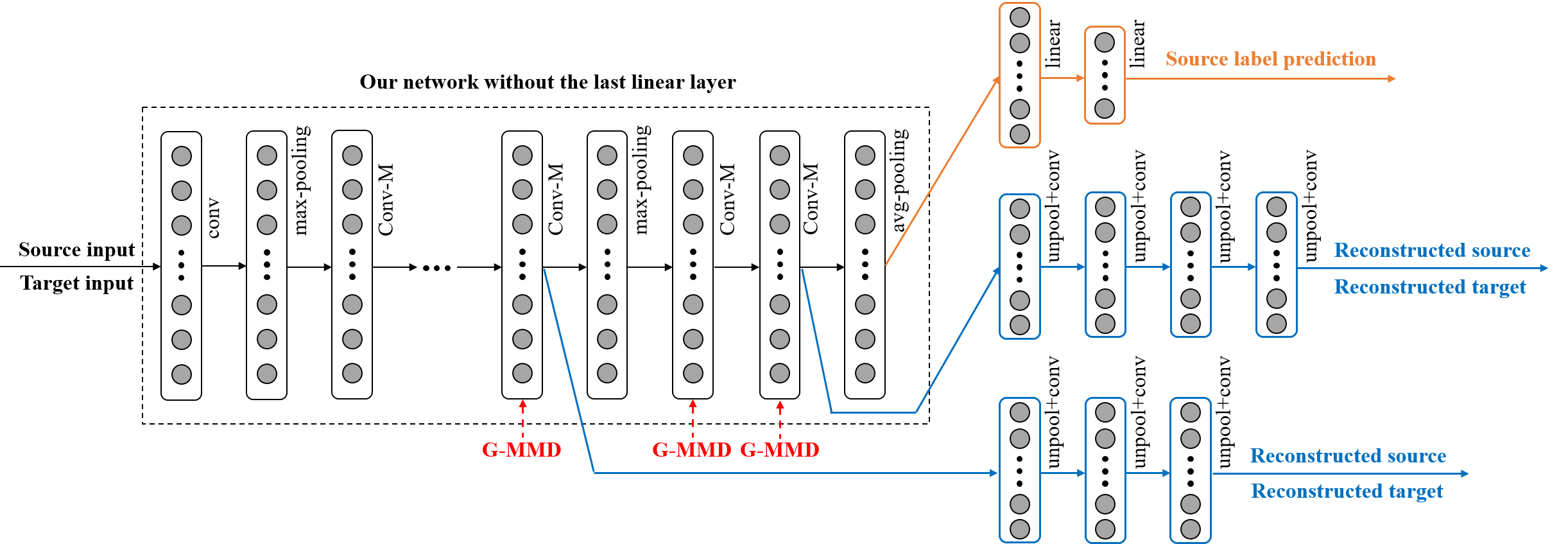}
	\caption{The unified framework of our DA method. The DNN simultaneously adapts feature representations (red and blue) and source label prediction (orange). The sampling ratio of target domain will be gradually increased during training.}
	\label{fig:da_arch}
\end{figure*}

%% file: Proposed_method.tex
\section{Proposed Method}
\label{sec:proposed_method}

Motivated by the observation described in Section~\ref{sec:introduction}, we propose a compact DNN architecture with a new basic module Conv-M. Our DA method gradually tunes the feature adaptation and label prediction.

\subsection{DNN Architecture with Conv-M}
\label{sec:our_dnn_arch}

Figure~\ref{fig:module_Conv-M} shows a Conv-M module used in our DNN.
According to the preliminary experiment and our analysis in Section~\ref{sec:introduction}, the design idea is to capture more diverse details at different levels, while using fewer parameters. 
To achieve this goal, the dilated convolution~\cite{Yu_ICLR16} for multi-resolution and deconvolution~\cite{Noh_ICCV13} are introduced. 
The dilated convolution can extract features with a larger receptive field without increasing the kernel size, \textit{e.g.}, extracting features from a 5$\times$5 window with a 3$\times$3 kernel.
The deconvolution is to reconstruct shapes of the input, providing distinct features from regular convolution.
In addition, to decrease redundant parameters, we implement the separable convolution inspired by separable wavelet filters~\cite{Sifre_CVPR13} for all types of convolution, including deconvolution, in Conv-M.

We visualize activations of convolution (middle) and deconvolution (right) in the same \textit{Conv-M} module in our network in Figure~\ref{fig:deconv_conv}.
Appearance details are extracted by convolution, while deconvolution tends to describe the completed shapes.
Therefore, the features extracted by convolution and deconvolution are \textit{complementary} so as to benefit DA.
In addition, the shapes captured by deconvolution are more \textit{generic} for a class of object compared to the appearance details extracted by convolution, which facilitates our DA strategy to explore divergence between classes for knowledge transfer.

The detailed design of Conv-M in Figure~\ref{fig:module_Conv-M} shows that
the input feature maps from the previous layer are respectively processed by regular convolution (\textit{conv}), dilated convolution (\textit{dilated conv}) and deconvolution (\textit{deconv}) in three branches.
Their outputs will be concatenated together. 
The pipelines of these three branches are: \textit{C1-C2-C3-dropout}, \textit{C4-DiC1-DiC2-dropout}, and \textit{C5-DeC1-DeC2-dropout}.
All of the three branches start with a 1$\times$1 convolution as linear projection. 
The parameters \textit{k} and \textit{s} are kernel size and stride. 
The dilation factor \textit{d} indicates that the receptive field is $(2^{d+1}-1)\times(2^{d+1}-1)$. 
The group number \textit{g} for separable convolution indicates that feature maps between two adjacent layers are separated into \textit{g} groups. 
The dropout ratio \textit{r} is fixed to 0.2. 
The output of deconvolution is cropped to its input size. 
ReLU is adopted for all nine convolutions, which is not shown in Figure~\ref{fig:module_Conv-M}.
The parameter number of Conv-M is computed as follows. Let $N_{P}$, $N_{C1}$, $N_{C2}$, $N_{C3}$, $N_{C4}$, $N_{DiC1}$, $N_{DiC2}$, $N_{C5}$, $N_{DeC1}$ and $N_{DeC2}$ denote the feature map numbers of \textit{C1}, \textit{C2}, \textit{C3}, \textit{C4}, \textit{DiC1}, \textit{DiC2}, \textit{C5}, \textit{DeC1} and \textit{DeC2}. The parameter number of the first branch in Conv-M is:
\begin{small}
	\begin{equation}
	\label{eq:conv-m_branch_1}
	N_{P}\cdot N_{C1}+\frac{N_{C1}\cdot N_{C2}\cdot k_{C2}^{2}}{g_{C2}}+\frac{N_{C2}\cdot N_{C3}\cdot k_{C3}^{2}}{g_{C3}}.
	\end{equation}
\end{small}
The parameter number of the second branch is:
\begin{small}
	\begin{equation}
	\label{eq:conv-m_branch_2}
	N_{P}\cdot N_{C4}+\frac{N_{C4}\cdot N_{DiC1}\cdot k_{DiC1}^{2}}{g_{DiC1}}+\frac{N_{DiC1}\cdot N_{DiC2}\cdot k_{DiC2}^{2}}{g_{DiC2}}.
	\end{equation}
\end{small}
The parameter number of the third branch is:
\begin{small}
	\begin{equation}
	\label{eq:conv-m_branch_3}
	N_{P}\cdot N_{C5}+\frac{N_{C5}\cdot N_{DeC1}\cdot k_{DeC1}^{2}}{g_{DeC1}}+\frac{N_{DeC1}\cdot N_{DeC2}\cdot k_{DeC2}^{2}}{g_{DeC2}}.
	\end{equation}
\end{small}

Our DNN architecture is shown in Table~\ref{table:our_dnn_arch}, which generally consists of \textit{convolution}, alternating \textit{max-pooling} and \textit{Conv-M}, \textit{avg-pooling} and \textit{linear}, as listed in the second column \textit{Types/Module}. 
Note that the last linear layer is for image classification only and will be removed when conducting DA tasks.
To fairly compare with other DA methods in Section~\ref{sec:experiments}, we include this layer into the estimation of total parameters as shown in the table.
The \textit{Output size} in the third column is multiplication of height, width and number of feature maps at each layer.
Specific parameters of a non Conv-M layer are listed in the fourth column \textit{Filter size/Stride}, while those of Conv-M are in the fifth column \textit{\#Feature maps (Conv-M)}. 
As the basic settings of Conv-M are represented in Figure~\ref{fig:module_Conv-M}, the fifth column only shows the feature map number of all nine convolutions: \textit{C1}, \textit{C2}, \textit{C3}, \textit{C4}, \textit{DiC1}, \textit{DiC2}, \textit{C5}, \textit{DeC1} and \textit{DeC2}.
For each of these nine convolutions, the feature map numbers between two max-pooling layers are same, and generally increased with the model depth.
The raw pixels of input images are processed by a regular convolution with a kernel size of 7$\times$7 which is much larger than the 1$\times$1 and 3$\times$3 kernels used in Conv-M.
Our preliminary experiment shows that for input image data, convolution with a smaller kernel (\textit{e.g.}, 3$\times$3) will degrade the classification accuracy by 1.5\%$\sim$2.5\%.
For Conv-M, on the other hand, using larger kernels (\textit{e.g.}, 5$\times$5) can only improve the performance by slightly 0.3\%$\sim$0.8\%. 
The final column \textit{\#Parameters} in Table~\ref{table:our_dnn_arch} lists the parameter numbers at each layer. Dominant parameter consumers are the two Conv-M modules (39\%) between the fourth max-pooling and the avg-pooling. 
The total number of parameters of our DNN is 4.1M. 

\subsection{Unsupervised Domain Alignment}
\label{sec:our_da_method}

Our DA method simultaneously adapts feature representations and source label prediction as shown in Figure~\ref{fig:da_arch}, given input data sampled from both source and target domains.
The sampling ratio of target domain will be gradually increased during training.
Formally, three terms are minimized in the unified framework: The reconstruction error of source and target samples (blue) for invariance learning, the discrepancy of hidden representations on layers between domains (red), and the prediction error of source labels (orange). 
For our DNN shown in Table~\ref{table:our_dnn_arch}, the last linear layer with 1000 neurons will be removed in DA tasks.
Extra layers, as shown in orange and blue in Figure~\ref{fig:da_arch}, are added during domain alignment training, while only the layers related to label prediction (orange) will be kept for testing.

\textbf{Invariance learning.} The error minimization of reconstructing input source and target samples is to force the DNN to learn more cross-domain features. 
The asymmetrical encoder-decoder architecture is adopted for sample reconstruction, as shown in Figure~\ref{fig:da_arch}. 
The encoder is our pre-trained DNN without the avg-pooling and last linear layers, while the decoder (blue) with fewer layers (compared to the encoder) consists of alternating un-pooling and regular convolution. 
The un-pooling in the decoder is to up-sample input feature maps using indexes obtained from the corresponding max-pooling layer in the encoder. 
The encoder is responsible for feature extraction, while the decoder is for restoring resolution. 
Our preliminary experiment shows that the asymmetrical structure only slightly decreases the final accuracy (averagely 0.4\%) but significantly accelerates the training speed, compared to symmetrical design. 
In addition, two decoders on different scales are introduced. 

\textbf{Representation discrepancy reduction.} 
Instead of using parametric criteria such as \textit{Kullback-Leibler divergence} to further reduce the cross-domain divergence, we adopt a non-parametric method to estimate the feature distribution distance between domains. Specifically, we minimize the maximum mean discrepancies (MMD) by Gretton \textit{et al.}~\cite{Gretton_NIPS06}. The MMD is defined as:
\begin{equation}
\label{eq:mmd_loss}
L_{M}=\bigg{\|}\frac{1}{N_{s}}\sum\limits_{1}^{N_{s}}\psi(x_{s})-\frac{1}{N_{t}}\sum\limits_{1}^{N_{t}}\psi(x_{t})\bigg{\|}_{\mathcal{H}}^{2},
\end{equation} 
where $x_{s}$ and $x_{t}$ are respectively input source and target, and $N_{s}$ and $N_{t}$ denote corresponding sample numbers. The function $\psi(\cdot)$ is a non-linear feature mapping. $\mathcal{H}$ is a universal reproducing kernel Hilbert space. The MMD criteria is denoted as G-MMD in our method, as we adopt the Gaussian kernel. As shown in Figure~\ref{fig:da_arch}, the G-MMD loss (red) is added to the last three Conv-M layers in our DNN.

\textbf{Source label prediction.} As shown in Figure~\ref{fig:da_arch}, we add two linear layers (orange), and the neuron numbers of the second one is specified for the dataset. No significant accuracy benefit is observed by adding linear layers more than two in our preliminary experiment.

%% file: Experiments.tex
\section{Experiments}
\label{sec:experiments}

Our DNN is trained on the benchmark dataset ImageNet'12~\cite{Russakovsky_IJCV15} and compared with well-known models on total parameter numbers and classification accuracy. 
Following the standard pipeline, we then fine-tune our trained model for unsupervised DA tasks on two popular datasets according to our DA method. The DA accuracy is compared with competitive methods.

\setlength{\tabcolsep}{9pt}
\begin{table}
	\begin{center}
		\caption{The comparison of our network and popular DNNs on ImageNet'12 classification accuracy and parameter numbers.}
		\label{table:exp_classification}
		\begin{threeparttable}
			\begin{tabular}{rccc}
				\hline\noalign{\smallskip}
				Method & \#Parameters & Top-1 & Top-5\\
				\hline\noalign{\smallskip}
				AlexNet~\cite{Alex_NIPS12}      & 61 M           & 57.2           & 80.3\\
				GoogLeNet~\cite{Szegedy_CVPR15} & 7 M            & 68.7           & 88.9\\
				VGG16~\cite{Simonyan_ICLR15}    & 134 M          & \textbf{71.9}  & \textbf{90.6}\\
				\hline
				\textbf{Our network}            & \textbf{4.1 M} & 68.9           & 89.0\\					
				\hline
			\end{tabular}
		\end{threeparttable}
		\vspace{-11pt}
	\end{center}
\end{table}
\setlength{\tabcolsep}{9pt}

\subsection{ImageNet Classification}
\label{sec:validation_classification}

We train our DNN on ImageNet'12 dataset, and set the parameters of our training solver according to the \textit{quick\_solver.prototxt} in Caffe~\cite{Jia_MM14}. 
The batch size is 64.
Table~\ref{table:exp_classification} compares the classification accuracy (\textit{Top-1}, \textit{Top-5}) and parameter numbers (\textit{\#Parameters}) of our DNN and AlexNet~\cite{Alex_NIPS12}, GoogLeNet~\cite{Szegedy_CVPR15}, and VGG16~\cite{Simonyan_ICLR15}. 
For AlexNet and GoogLeNet, we directly use the trained models provided by Caffe. 
The VGG16's result is obtained from the original paper~\cite{Simonyan_ICLR15}. 
Our DNN achieves GoogLeNet-level accuracy, while the total parameter numbers (4.1M) is only 59\% of GoogLeNet.

\setlength{\tabcolsep}{8pt}
\begin{table*}
	\begin{center}
		\caption{Unsupervised DA accuracy of our method and previous algorithms on Office-31 dataset.}
		\label{table:exp_office-31}
		\vspace{-4pt}
		\small
		\begin{threeparttable}
			\begin{tabular}{l|c|cccccc}
				\hline\noalign{\smallskip}
				Method & \#Parameters\tnote{1} & A$\rightarrow$W & D$\rightarrow$W & W$\rightarrow$D & W$\rightarrow$A & A$\rightarrow$D & D$\rightarrow$A\\
				\hline\hline\noalign{\smallskip}
				GFK~\cite{Gong_ICML13}            & -      & 39.8 & 79.1 & 74.6 & 37.1 & 37.9 & 37.9\\	
			    SA~\cite{Fernando_ICCV13}         & -      & 45.0 & 64.8 & 69.9 & 39.3 & 38.8 & 42.0\\
			    DLID~\cite{Chopra_ICMLW13}        & -      & 51.9 & 78.2 & 89.9 & -    & -    & -   \\
			    DDC~\cite{Tzeng_arxiv14}          & -      & 61.8 & 95.0 & 98.5 & 52.2 & 64.4 & 52.1\\
			    DAN~\cite{Long_ICML15}            & 61 M & 68.5 & 96.0 & 99.0 & 53.1 & 67.0 & 54.0\\
			    GRL~\cite{Ganin_ICML15}      & 61 M & 73.0 & 96.4 & 99.2 & 53.6 & 72.8 & 54.4\\
			    TRANSDUCTION~\cite{Sener_NIPS16}  & 61 M & 80.4 & 96.2 & 98.9 & 62.5 & \textbf{83.9} & 56.7\\
			    \hline
			    GRL (Rev-FaConvNet)      &	 4.8 M           & 74.1 & 96.5 & 99.2 & 54.3 &　73.4 & 55.3\\
			    Our DA (Rev-FaConvNet)   & 4.8 M           & 77.0 & 96.5 & 99.2 & 58.4 & 75.9 & 58.1\\		             	      			
                \hline          
                GRL (Our net)            & \textbf{4.1 M}  & 80.1 & 96.7 & 99.2 & 64.1 & 78.0 & 65.4\\               
				\textbf{Our DA (Our net)}   & \textbf{4.1 M}      & \textbf{82.6} & \textbf{97.0} & \textbf{99.4} 
				                                                  & \textbf{67.4} & 80.1 & \textbf{67.3}\\				
				\hline\hline
				Baseline: Our DA (GoogLeNet)                     & 7 M   & 83.0 & 96.9 & 99.5 & 67.7 & 80.5 & 67.5\\
				Baseline: Our DA (FaConvNet)                     & 2.8 M & 73.9 & 96.3 & 99.1 & 54.1 & 73.2 & 55.2\\
				\hline
			\end{tabular}
			\begin{tablenotes}
				\footnotesize
				\item[1]  
				Most of methods will remove the last linear layer of a pre-trained network, and add extra layers for DA. According to Section~\ref{sec:validation_da}, our DNN will be smaller after the change. The size of other models will also be slightly different, but the actual size is not reported in ~\cite{Long_ICML15,Sener_NIPS16}. We hence directly report the total parameter numbers of the pre-trained network for fair comparison.
			\end{tablenotes}		
		\end{threeparttable}
		\vspace{-8pt}
	\end{center}
\end{table*}
\setlength{\tabcolsep}{8pt}

\setlength{\tabcolsep}{3pt}
\begin{table*}
	\begin{center}
		\caption{Unsupervised DA accuracy of our method and previous algorithms on Office-Caltech dataset.}
		\label{table:exp_office-caltech}
	    \small
		\begin{threeparttable}
			\begin{tabular}{l|c|cccccccccccc}
				\hline\noalign{\smallskip}
				Method & \#Param.\tnote{1} & A$\rightarrow$W & D$\rightarrow$W　& W$\rightarrow$D　& A$\rightarrow$D　
				& D$\rightarrow$A　& W$\rightarrow$A　& A$\rightarrow$C　& W$\rightarrow$C　& D$\rightarrow$C　
				& C$\rightarrow$A　& C$\rightarrow$W　& C$\rightarrow$D\\
				\hline\hline\noalign{\smallskip}
				TCA~\cite{Pan_TNNLS11}            & -      & 84.4 & 96.9 & 99.4 & 82.8 & 90.4 & 85.6 & 81.2 
				& 75.5 & 79.6 & 92.1 & 88.1 & 87.9\\
				GFK~\cite{Gong_ICML13}            & -      & 89.5 & 97.0 & 98.1 & 86.0 & 89.8 & 88.5 & 76.2
				& 77.1 & 77.9 & 90.7 & 78.0 & 77.1\\
				DDC~\cite{Tzeng_arxiv14}          & -      & 86.1 & 98.2 & \textbf{100.0} & 89.0 & 89.5 & 84.9
				& 85.0 & 78.0 & 81.1 & 91.9 & 85.4 & 88.8\\				                                           
				DAN~\cite{Long_ICML15}            & 61 M   & 93.8 & 99.0 & \textbf{100.0} & 92.4 & 92.0 & 92.1
				& 85.1 & 84.3 & 82.4 & 92.0 & 90.6 & 90.5\\	                                           
				RTN~\cite{Long_NIPS16}            & 61 M & \textbf{97.0} & 98.8 & \textbf{100.0} 
				& 94.6 & 95.5 & 93.1 & 88.5 & 88.4 & 84.3 & 94.4 
				& 96.6 & 92.9\\			
				\hline
				DAN (Rev-FaConvNet)    & 4.8 M        & 94.0 & 99.1 & \textbf{100.0} & 92.7 & 92.3 & 92.2 & 85.5 & 84.6 
				                                      & 82.6 & 92.3 & 90.9 & 90.8\\
				Our DA (Rev-FaConvNet) & 4.8 M        & 94.9 & 99.2 & \textbf{100.0} & 93.3 & 93.3 & 92.5 & 86.5 & 85.9
				                                      & 83.1 & 93.0 & 93.0 & 91.5\\
				\hline               
				DAN (Our net)        & \textbf{4.1 M} & 95.0 & 99.2 & \textbf{100.0} & 96.0 & 94.8 & 95.2 & 91.6 
				                                      & 90.4 & 90.7 & 94.4 & 95.0    & 94.3             
				                                      \\                                   
				\textbf{Our DA (Our net)}    & \textbf{4.1 M} & 95.6 & \textbf{99.7} & \textbf{100.0} & \textbf{96.8}
				                                     & \textbf{96.0} & \textbf{95.6} & \textbf{92.5} 
				                                     & \textbf{91.6} & \textbf{91.4} & \textbf{95.3}
				                                     & \textbf{97.2} & \textbf{95.3}\\
				\hline\hline
				Baseline: Our DA (GoogLeNet) & 7 M       & 95.9 & 99.7 & 100.0 & 97.1 & 96.2 & 95.9 & 92.9 
				                                     & 92.0 & 91.5 & 95.6  & 97.4 & 95.7\\
				Baseline: Our DA (FaConvNet) & 2.8 M     & 94.5 & 99.1 & 99.8  & 92.0 & 91.8 & 91.0 & 83.7 & 83.4 & 80.1
				                                     & 92.8 & 91.1 & 89.8\\	
				\hline
			\end{tabular}
			\begin{tablenotes}
				\footnotesize
				\item[1]  
				Please see the footnote of Table~\ref{table:exp_office-31} for the explanation of parameter numbers. 
			\end{tablenotes}	
		\end{threeparttable}
		\vspace{-15pt}
	\end{center}
\end{table*}
\setlength{\tabcolsep}{3pt}

\setlength{\tabcolsep}{5pt}
\begin{table*}
	\begin{center}
		\caption{Contribution of non-regular convolution in our Conv-M module on Office-31 dataset.}
		\label{table:sensitivity_convm}
		\vspace{-4pt}
		\small
		\begin{threeparttable}
			\begin{tabular}{l|c|c|cccccc}
				\hline\noalign{\smallskip}
				& \#Parameter & Classification & A$\rightarrow$W & D$\rightarrow$W & W$\rightarrow$D & W$\rightarrow$A & A$\rightarrow$D & D$\rightarrow$A\\
				\hline\noalign{\smallskip}
				Our DA (Our net1)            & 4.1 M    & 62.2    & 74.2          & 96.5          & 99.2
				                                                  & 56.2          & 74.1          & 56.0\\		
				Our DA (Our net)             & 4.1 M    & 68.9    & \textbf{82.6} & \textbf{97.0} & \textbf{99.4} 
				                                                  & \textbf{67.4} & \textbf{80.1} & \textbf{67.3}\\	
				\hline
			\end{tabular}
		\end{threeparttable}
		\vspace{-15pt}
	\end{center}
\end{table*}
\setlength{\tabcolsep}{5pt}

\subsection{Unsupervised DA}
\label{sec:validation_da}

\textbf{Office-31}. This standard benchmark consists of 4,652 images of 31 categories collected from three distinct domains ~\cite{Saenko_ECCV10}: \textit{AMAZON} (A), \textit{WEBCAM} (W) and \textit{DSLR} (D). The samples of these three domains are respectively downloaded from \textit{amazon.com}, taken by web camera and taken by digital SLR camera in an office environment with different photographic settings. All six DA tasks between the three domains will be adopted for completeness: A$\rightarrow$W, D$\rightarrow$W, W$\rightarrow$D, W$\rightarrow$A, A$\rightarrow$D and D$\rightarrow$A. 

\textbf{Office-Caltech.} It is a popular dataset~\cite{Gong_CVPR12} composed of 10 overlapping categories from the Office-31 and Caltech-256 (C)~\cite{Griffin_Caltech07} datasets. All twelve DA tasks are used: A$\rightarrow$W, D$\rightarrow$W, W$\rightarrow$D, A$\rightarrow$D, D$\rightarrow$A, W$\rightarrow$A, A$\rightarrow$C, W$\rightarrow$C, D$\rightarrow$C, C$\rightarrow$A, C$\rightarrow$W and C$\rightarrow$D. The Office-31 dataset is more challenging as it has more categories of images, while Office-Caltech provides more DA tasks to observe the dataset bias~\cite{Torralba_CVPR11}. 

\textbf{Methods.} 
We compare our method with the nine previous competitive DA methods: TCA~\cite{Pan_TNNLS11}, GFK~\cite{Gong_ICML13}, SA~\cite{Fernando_ICCV13}, DLID~\cite{Chopra_ICMLW13}, DDC~\cite{Tzeng_arxiv14}, DAN~\cite{Long_ICML15}, GRL~\cite{Ganin_ICML15}, TRANSDUCTION~\cite{Sener_NIPS16} and RTN~\cite{Long_NIPS16}.
TCA and GFK are conventional methods, while the others are DNN based.

\textbf{Networks.}
Five DNNs are used in our experiments: AlexNet (61M), Rev-FaConvNet (4.8M), our DNN (4.1M), GoogLeNet (7M) and FaConvNet (2.8M).
DA methods DAN, GRL, TRANSDUCTION and RTN originally use pre-trained AlexNet, according to their papers.
Rev-FaConvNet achieves much better DA accuracy compared to SqueezeNet, Rev-SqueezeNet and FaConvNet as shown in our preliminary experiments in Table~\ref{table:comp_classification_da}.
FaConvNet, Rev-FaConvNet and our DNN all reach GoogLeNet-level classification accuracy.
In this work, we use GoogLeNet and FaConvNet as baselines for comparison.

\textbf{Experiments.}
Besides running previous DA methods on AlexNet, we also run the following eight experiments to quantize the contribution of our DNN and our DA method:\\
(1) GRL (Rev-FaConvNet): Running GRL on Rev-FaConvNet;\\
(2) GRL (Our net): Running GRL on our DNN;\\
(3) DAN (Rev-FaConvNet): Running DAN on Rev-FaConvNet;\\
(4) DAN (Our net): Running DAN on our DNN;\\
(5) Our DA (Rev-FaConvNet): Running our DA method on Rev-FaConvNet;\\
(6) Our DA (FaConvNet): Running our DA method on FaConvNet, and the result is used as a baseline;\\
(7) Our DA (GoogLeNet): Running our DA method on GoogLeNet, and the result is used as a baseline;\\
(8) Our DA (Our net): Running our DA method on our DNN, and this is \textbf{our final result}.

\textbf{Parameter settings.} 
We follow the specific description of all previous DA methods in their papers. 
The hyper-parameter of SA is selected based on cross-validation, which is consistent with other papers~\cite{Ganin_ICML15,Sener_NIPS16}.
For our DA method that is based on our pre-trained network on ImageNet'12, the \textit{convolution} and the first three \textit{Conv-M} shown in Table~\ref{table:our_dnn_arch} are frozen, as the Office-31 and Office-Caltech datasets are rather small-scale.
For all newly added layers as shown in orange and blue in Figure~\ref{fig:da_arch} which are trained from scratch, their learning rate is ten times higher. 
The learning rate policy we adopt is \textit{poly} as described in Caffe, and the initial value is 0.0009 with the power fixed to 0.5.
The batch size is 64, and the sampling ratio of target domains is uniformly increased from 30\% to 70\% during training.
In the testing stage, the new layers for sample reconstruction are removed, as aforementioned in Section~\ref{sec:our_da_method}.
For the remaining new layers for label prediction (orange) in Figure~\ref{fig:da_arch}, the neuron numbers of the first linear layer is 256, while those of the second one is 31 for Office-31 dataset and 10 for Office-Caltech dataset.
The G-MMD loss is added to the last three Conv-M layers of our DNN.
The regularization hyper-parameter of G-MMD loss is fixed to 0.3 across all datasets, and the bandwidth of the Gaussian kernel is the median pairwise distance~\cite{Gretton_NIPS12} on training set.

Based on NVIDIA GTX TITAN X, the inference speed of SqueezeNet and Rev-SqueezeNet is faster than that of FaConvNet, Rev-FaConvNet and our network, though they cannot obtain GoogLeNet-level classification and DA. 
Specifically, Rev-SqueezeNet is 22\% slower than that of SqueezeNet, and Rev-FaConvNet decreases the speed of FaConvNet by 12\%.
Our network consumes 11\% less time compared to FaConvNet.  

Table~\ref{table:exp_office-31} and Table~\ref{table:exp_office-caltech} respectively summarize the DA accuracy on Office-31 and Office-Caltech datasets.
Both tables are separated into four groups by rows.
The first group is the previous DA methods based on AlexNet.
The second group compares previous and our DA methods on Rev-FaConvNet, while the third group compares DA methods on our DNN.
The fourth group provides result of our DA method on GoogLeNet and FaConvNet as baselines.
The results in the two tables are analyzed from the following three aspects:

First, our DNN approaches GoogLeNet's DA accuracy on the same DA method, while the gap between GoogLeNet and previous compact DNNs (FaConvNet and Rev-FaConvNet) is much larger, according to the four observations: \textit{Our DA (Our net)}, \textit{Our DA (GoogLeNet)}, \textit{Our DA (FaConvNet)} and \textit{Our DA (Rev-FaConvNet)} in Table~\ref{table:exp_office-31} and Table~\ref{table:exp_office-caltech}. 
Though FaConvNet, Rev-FaConvNet and our DNN all obtain GoogLeNet-Level classification accuracy, only our DNN has matched accuracy on both classification and DA. 
Moreover, our DNN (4.1M) is smaller than Rev-FaConvNet (4.8M).
Our DNN also outperforms AlexNet using the same DA method, as the comparison of \textit{GRL} and \textit{GRL (Our net)} in Table~\ref{table:exp_office-31} shows.

Second, our DA method outperforms GRL and DAN, based on the same DNN, according to the four comparisons: \textit{GRL (Rev-FaConvNet)} and \textit{Our DA (Rev-FaConvNet)} in Table~\ref{table:exp_office-31}, \textit{GRL (Our net)} and \textit{Our DA (Our net)} in Table~\ref{table:exp_office-31}, \textit{DAN (Rev-FaConvNet)} and \textit{Our DA (Rev-FaConvNet)} in Table~\ref{table:exp_office-caltech}, and \textit{DAN (Our net)} and \textit{Our DA (Our net)} in Table~\ref{table:exp_office-caltech}.

Third, put all together, our DA method based on our DNN achieves state-of-the-art on sixteen of total eighteen DA tasks on two datasets, as shown on the last row of these two tables (\textit{Our DA (Our net)}).
The other two is \textit{A}$\rightarrow$\textit{D} in Table~\ref{table:exp_office-31} and \textit{A}$\rightarrow$\textit{W} in Table~\ref{table:exp_office-caltech}. 
We boost the accuracy of task \textit{D}$\rightarrow$\textit{A} by 10.6\% compared to TRANSDUCTION, as shown in Table~\ref{table:exp_office-31}.
On Office-31 dataset, the accuracy gap between the tasks \textit{D}$\rightarrow$\textit{W} and \textit{W}$\rightarrow$\textit{D} is 2.4\%, while the gap between \textit{A}$\rightarrow$\textit{W} and \textit{W}$\rightarrow$\textit{A} greatly increases to 15.2\%, indicating larger appearance difference between domains A and W. 
The domain difference between A and D is also larger than that between D and W. 
In other words, on Office-31 dataset, transfer (in two directions) between D and W is relatively easier for our DA method, while other two are more difficult, which is consistent with the results from previous DA methods.
On Office-Caltech dataset, the bilateral transfer between C and W gets the largest accuracy gap (5.6\%) in our DA method, as shown in Table~\ref{table:exp_office-caltech}.

\setlength{\tabcolsep}{3pt}
\begin{table}
	\begin{center}
		\caption{DA accuracy of our method without including specified component on Office-31 dataset.}
		\label{table:sensitivity_office-31}
		\vspace{-4pt}
		\small
		\begin{threeparttable}
			\begin{tabular}{c|ccccccc}
				\hline\noalign{\smallskip}
				Method & A$\rightarrow$W & D$\rightarrow$W & W$\rightarrow$D & W$\rightarrow$A & A$\rightarrow$D & D$\rightarrow$A\\
				\hline\noalign{\smallskip}
				No G-MMD  & 76.7 & 96.5 & 99.2 & 62.0 & 77.5 & 64.7\\
				No recons.& 79.6 & 95.4 & 99.3 & 64.4 & 77.3 & 62.1\\
				\hline
				\textbf{All} & \textbf{82.6} & \textbf{97.0} & \textbf{99.4} & \textbf{67.4} & \textbf{80.1} & \textbf{67.3}\\				
				\hline
			\end{tabular}		
		\end{threeparttable}
		\vspace{-8pt}
	\end{center}
\end{table}
\setlength{\tabcolsep}{3pt}

\setlength{\tabcolsep}{3pt}
\begin{table}
	\begin{center}
		\caption{DA accuracy of our method without including specified component on Office-Caltech dataset.}
		\label{table:sensitivity_office-caltech}
		\vspace{-4pt}
		\small
		\begin{threeparttable}
			\begin{tabular}{c|ccccccc}
				\hline\noalign{\smallskip}
				Method & A$\rightarrow$W & D$\rightarrow$W & A$\rightarrow$D & A$\rightarrow$C & W$\rightarrow$C & D$\rightarrow$C\\
				\hline\noalign{\smallskip}
				No G-MMD  & 91.1 & 99.6 & 93.4 & 90.9 & 87.1 & 87.8\\
				No recons.& 93.9 & 99.4 & 95.0 & 88.7 & 89.8 & 86.6\\
				\hline
				\textbf{All} & \textbf{95.6} & \textbf{99.7} & \textbf{96.8} & \textbf{92.5} & \textbf{91.6} & \textbf{91.4}\\				
				\hline
			\end{tabular}		
		\end{threeparttable}
		\vspace{-15pt}
	\end{center}
\end{table}
\setlength{\tabcolsep}{3pt}

\subsection{Sensitivity Analysis}
\label{sec:sensitivity}

\textbf{Convolution in Conv-M.} To validate the contribution of non-regular convolution (dilated convolution and improved deconvolution) in our Conv-M module, we replace all non-regular convolution with regular ones and keep the 3$\times$3 kernel size unchanged. The first row \textit{Our DA (Our net1)} in Table~\ref{table:sensitivity_convm} shows the result, and the second row \textit{Our DA (Our net)} is our original solution. Significant accuracy drop can be observed on classification and almost all DA tasks. The comparison in Table~\ref{table:sensitivity_convm} indicates the importance of features extracted by dilated convolution and improved deconvolution in our Conv-M. 

\textbf{Reconstrution and G-MMD.} Based on our DNN, Table~\ref{table:sensitivity_office-31} and Table~\ref{table:sensitivity_office-caltech} respectively show the contribution of two components of our DA methods (sample reconstruction and G-MMD) on Office-31 and Office-Caltech datasets. The row \textit{No G-MMD} in two tables shows the result obtained by removing G-MMD from our DA method, while the row \textit{No recons.} corresponds to our method without including sample reconstruction. For these two rows, lower accuracy indicates more contribution of the component. The row \textit{All} is the regular result without removing any component, which is the same as the respective row \textit{Our DA (Our net)} in Table~\ref{table:exp_office-31} and Table~\ref{table:exp_office-caltech}. For Office-31 dataset shown in Table~\ref{table:sensitivity_office-31}, reconstruction is more important for the transfers \textit{D}$\rightarrow$\textit{W} and \textit{D}$\rightarrow$\textit{A}, while \textit{A}$\rightarrow$\textit{W} and \textit{W}$\rightarrow$\textit{A} rely more on G-MMD. Table~\ref{table:sensitivity_office-caltech} demonstrates that the contributions of reconstruction and G-MMD are almost the same.

%% file: Conclusion.tex
\section{Conclusion}
\label{sec:conclusion}
In this paper, we present a compact DNN architecture and unsupervised DA method, based on our observation that current small DNNs (SqueezeNet and FaConvNet) have unmatched accuracy on classification and DA, \textit{e.g.}, a DNN with GoogLeNet-level classification accuracy only obtains AlexNet-level DA accuracy.
The basic module used in our DNN, Conv-M, introduces multi-scale convolution and deconvolution without using kernels larger than 3$\times$3.
The unified framework of our DA method learns cross-domain features by sample reconstruction and G-MMD, and simultaneously tunes label prediction.
The parameter numbers of our DNN is only 59\% of GoogLeNet, while experiments show that our DNN obtains GoogLeNet-level accuracy both on classification and DA.
Our DA method slightly outperforms previous competitive GRL and DA.
In addition, our method based on our DNN achieves state-of-the-art on sixteen of total eighteen DA tasks on the popular Office-31 and Office-Caltech datasets.\newline

\noindent\textbf{Acknowledgments.} This work is in part supported by NSF CCF-1615475 and DOE SC0017030. Any opinions, findings and conclusions or recommendations expressed in this material are those of the authors and do not necessarily reflect the views of grant agencies or their contractors.